\documentclass[sigconf]{acmart}
\setcopyright{acmlicensed}
\acmYear{2026}
\acmDOI{}
\acmConference[LAK '26]{16th International Learning Analytics and Knowledge Conference}{April 2026}{Online}
\acmBooktitle{Proceedings of the 16th International Learning Analytics and Knowledge Conference (LAK '26)}
\usepackage{booktabs}
\usepackage{graphicx}
\usepackage{tabularx}
\usepackage{caption}
\usepackage{subcaption}
\usepackage{xurl}
\usepackage{lipsum}

\begin{document}
\title[Modeling Grammatical Hypothesis Testing in Young Learners]{Modeling Grammatical Hypothesis Testing in Young Learners: A Sequence-Based Learning Analytics Study of Morphosyntactic Reasoning in an Interactive Game}


\author{Thierry Geoffre}
\orcid{0000-0003-0237-2310}
\affiliation{\institution{University of Luxembourg}
             \department{FHSE - Department of Education and Social Work}
             \city{Esch-sur-Alzette}
             \country{Luxembourg}}
\email{thierry.geoffre@uni.lu}

\author{Trystan Geoffre}
\orcid{0009-0006-2481-0982}
\affiliation{\institution{Universit\'e de Fribourg}
             \department{Institut de Plurilinguisme}
             \city{Fribourg}
             \country{Switzerland}}
\email{trystan.geoffre@unifr.ch}

\renewcommand{\shortauthors}{Geoffre and Geoffre}

\begin{abstract}
This study investigates grammatical reasoning in primary school learners through a sequence-based learning analytics approach, leveraging fine-grained action sequences from an interactive game targeting morphosyntactic agreement in French. Unlike traditional assessments that rely on final answers, we treat each slider movement as a hypothesis-testing action, capturing real-time cognitive strategies during sentence construction. Analyzing 597 gameplay sessions (9,783 actions) from 100 students aged 8--11 in authentic classroom settings, we introduce Hamming distance to quantify proximity to valid grammatical solutions and examine convergence patterns across exercises with varying levels of difficulty. Results reveal that determiners and verbs are key sites of difficulty, with action sequences deviating from left-to-right usual treatment. This suggests learners often fix the verb first and adjust preceding elements. Exercises with fewer solutions exhibit slower and more erratic convergence, while changes in the closest valid solution indicate dynamic hypothesis revision. Our findings demonstrate how sequence-based analytics can uncover hidden dimensions of linguistic reasoning, offering a foundation for real-time scaffolding and teacher-facing tools in linguistically diverse classrooms.
\end{abstract}


\maketitle

\section{Introduction}
Reading comprehension is a fundamental skill that enables learning in all subjects and full participation in society. However, in increasingly heterogeneous primary classrooms, such as in the multilingual contexts of French-speaking Switzerland and Luxembourg, many students struggle to master the written schooling language due to linguistic diversity, developmental language disorders, or limited prior exposure. Traditional approaches with the same text or the same activity for all students often fail to address these varied needs, creating a pressing demand for inclusivity and personalized learning pathways.

Adaptive learning could be an answer, making use of the opportunities offered by digital technology and relying on data-driven interventions. With this in mind, we developed the game-based platform GamesHub\cite{aebischer2025}, with an adaptable learner's interface and a teacher's interface providing data and feedback. The tool implements an adaptive learning engine that dynamically personalizes each student's pathway based on real-time performance, while generating rich interaction logs and sequences of actions.

The platform delivers short, game-like activities that target the key micro- and integrative processes of comprehension \cite{irwin2007, giasson2008}, the subskills that allow words' and groups of words' recognition, and the identification of links between pieces of information in short texts, such as the understanding of temporal, spatial or logical links between two explicit pieces of information; the generation of inferences (understanding implicit information); the resolution of anaphora (for example, the use of a pronoun referring back to a word used earlier in the text, to avoid repetition), or the interpretation of morphosyntactic cues.

In French, morphograms (grammatical marks) are crucial as they are very often silent \cite{karmiloff1981, pacton2017, brissaud2018}: for example, the four flexions of the adjective \textit{bleu} (blue) will spell differently (\textit{bleu; bleue; bleus; bleues}) depending on gender and number combinations, but will sound strictly the same. Then, in the noun phrases \textit{le vélo bleu} (the blue bicycle), \textit{les vélos bleus} (the blue bicycles), \textit{la bouteille bleue} (the blue bottle), \textit{les bouteilles bleues} (the blue bottles), gender and number of the adjective are clearly visible but always silent. This creates deep difficulties in writing French but also huge challenges to foster reading comprehension as the visual identification of a morphogram may give a real help in understanding. For example, in the phrases \textit{je suis habillé / je suis habillée} (I am dressed) that sound the same, the presence of the final morphogram -e indicates if it is masculine or feminine. The same can be observed with verbs: in the phrases \textit{il marche / ils marchent} (he walks / they walk) that sound the same, the absence or presence of the morphograms \textit{-s} and \textit{-nt} allows to understand a singular or a plural sentence when the oral is unable to distinguish the meaning. The fact that some words will be linked by gender, number or person morphograms defines the morphosyntactic structure of the French language. These morphograms are visual cues of understanding. And we call "chain of agreement" the string of N words that are linked by morphosyntax.

We have developed a game that specifically targets the interpretation of these morphosyntactic cues and the grammatical reasoning that must be mobilized. This game is locally named "Tirettes" (French for sliders or zippers), and belongs to the online app \textit{L'Ortho\-dyss\'ee des Gram} \cite{geoffre2026}. This game asks the player to build a grammatically correct sentence by moving sliders. Each slider wears several words or several spelling of the same word (e.g. the masculine/feminine + singular/plural flexions of an adjective). These sliders are arranged side by side, forming the syntactic structure of a sentence. By moving them, it is possible to align the words to respect the chains of agreement and produce a grammatically correct sentence. Then the student may validate the sentence to receive feedback. This interaction between the player and the sliders can be recorded and make the strategies of the player visible. Each slider move is not a "response", it is an action reflecting a hypothesis about linguistic structure. A child moving an adjective from \textit{petit} to \textit{petite} ("small" from masculine to feminine form) is not just selecting a word, he or she is testing a hypothesis about gender agreement: a vertical move impacts the agreement chain on the morphosyntactic (horizontal) axis (see 2.2). This is a step forward accessing the cognitive process of applying morphosyntactic rules to construct coherent utterances which is rarely observable in traditional classroom settings. Learners produce sentences, and educators judge them correct or incorrect. But how do children reason through agreement rules? Do they apply a rule? Guess? Test hypotheses? Revert? \textit{Tirettes} makes these strategies visible in a global activity of reading the sentence. These strategies, linked to the gameplay and to the understanding of the morphosyntax of French, may be later transferred into reasoning when reading and writing.

\subsection*{Our Contribution}
This paper presents a first step in the data-driven analysis of the use of the Tirettes game. We study grammatical reasoning by treating sequences of actions as the fundamental data, and not only the final answers. We analyze 597 learner sessions (9,783 actions) to answer: \textit{What patterns in action sequences reveal about the children grammatical reasoning when playing an interactive grammatical game?}

\section{Related Work}
Our work intersects three interrelated domains of research: (1) learning analytics approaches on (2) research on development of the morphographic competency in French, based on (3) game-based assessment and interaction logging for capturing strategies that model \textit{process} rather than \textit{product} in educational tasks.

\subsection{Development of the Morphographic Competency in French}
In French, grammatical agreement relies heavily on orthographic cues that are often phonologically silent, making crucial their identification (reading comprehension) and production (writing). The learning of the morphosyntax of French is then a challenge particularly acute for learners of K12 education. While classroom-based studies have documented children's errors in agreement marking \cite{fayol2016}, they rarely capture the \textit{dynamic process} of rule application or hypothesis testing. Digital tools offer a promising avenue to bridge this gap. Nevertheless, most existing studies focus on production or judgment tasks, not on the \textit{exploratory reasoning} that precedes a final validation. Moreover, a recent extensive review of digital tools promising to target grammar and grammatical reasoning in French has shown the actual uselessness of current proposals \cite{arseneau2023}. Works in educational linguistics has begun to integrate log data to study how learners navigate morphological complexity \cite{ravid2002} but French (with its high degree of homophony in inflectional forms) remains underrepresented. Our study addresses this gap by analyzing how children navigate agreement constraints through iterative, low-stakes interactions in a game-based environment.

\subsection{Game-Based Assessment and the Visibility of Cognitive Strategies}
Serious games and interactive learning environments offer unique affordances for capturing implicit reasoning through observable actions \cite{shute2013}. Unlike multiple-choice or fill-in-the-blank formats, game mechanics can be designed so that each interaction reflects a hypothesis or decision about underlying rules. This principle underpins \textit{stealth assessment} \cite{shute2011}, where evidence of competence is gathered unobtrusively during gameplay. In literacy, games like \textit{GraphoGame} have logged keystrokes to study decoding strategies \cite{ojanen2015}, while other platforms have used drag-and-drop or sequencing actions to infer syntactic awareness. However, few studies have leveraged \textit{slider-based mechanics} where multiple linguistic variants are presented simultaneously to make grammatical agreement decisions visible as exploratory actions. This refers to the fact that the spelling of a word in a sentence is chosen among different flexions (the vertical slider, or paradigmatic axis) according to the syntactic context (the horizontal line of the sentence, or morphosyntactic axis). The \textit{Tirettes} game embodies this work by transforming abstract morphosyntactic rules into manipulable objects, thereby externalizing the learner's reasoning process in real time.

\subsection{From Final Answers to Action Sequences in Learning Analytics}
Traditional educational assessment often treats learners' responses as static outcomes (correct \textit{vs} incorrect) without considering the cognitive journey that led to them. In contrast, a growing body of learning analytics research emphasizes the value of \textit{process data}: sequences of interactions, navigation paths, or editing behaviors that reveal how learners engage with tasks \cite{bechet2012}. For instance, in programming education, action sequences in code editors have been used to infer problem-solving strategies; in mathematics, step-level log data has enabled the detection of misconceptions beyond final answers. More recently, sequence mining and process mining techniques have been applied to uncover recurrent behavioral patterns in digital learning environments \cite{bannert2014, fatahi2018}. Our work aligns with this shift toward \textit{process-oriented analytics}, but applies it to a domain underexplored in LAK: early language development, specifically morphosyntactic reasoning in written French.

\subsection{Synthesis and Gap}
Collectively, these strands highlight a clear trend: the move from outcome-based to process-based analytics, the potential of game mechanics to make visible actions and strategies, and the need for tools to support linguistically diverse learners. Yet, to our knowledge, no study has combined these perspectives to analyze \textit{grammatical reasoning as a sequence of hypothesis-testing actions} in a real-world inclusive primary education context. Our work contributes to LAK by demonstrating how rich interaction logs from an adaptive game can reveal otherwise invisible reasoning strategies; offering both theoretical insights into morphosyntactic development and practical pathways for data-informed personalization in heterogeneous classrooms.

\section{Methodology}
The GamesHub platform has been implemented in 11 primary classrooms in French speaking Switzerland (N=200 students, 3rd to 5th grades, ages 8-11) during 10 weeks between January and May 2025 (2 sessions per week), in connection with the project DOCTA\textsuperscript{2}LE-FR\footnote{Project Documenter et Tester l'Apprentissage Adaptatif à L'École en Français, DOCTA\textsuperscript{2}LE-FR, funded by the Swiss National Science Foundation SNSF, CHF 709,439, grant n°100019\_215373, 48 months funding}. This project, funded by the Swiss National Science Foundation, aims among others to test and document the use of adaptive learning in primary school, in the context of the sub-skills implied in the reading-comprehension of short texts. Then, the use of the platform offers a unique opportunity to investigate how adaptive game-based learning unfolds in authentic settings.

\subsection{Platform and Game: Tirettes}
Ten games were available, and learners' personalized pathways could rely on more than 1200 game levels ("exercises" in the following to avoid confusion with level of difficulty). Among these ten games, we decided to start our analyses with data from the Tirettes game as the sequences of actions are really linked to reasoning.

\textit{Tirettes}\footnote{\url{https://files.atypon.com/acm/92676739cffed56ac707f1391545c41e}} asks players to build grammatically correct sentences by moving sliders. Each slider contains several words or different inflections of the same word. By moving sliders, players align words to respect agreement chains and produce correct sentences.

\begin{figure}[h]
\centering
\begin{minipage}{0.48\textwidth}
\centering
\includegraphics[width=\textwidth]{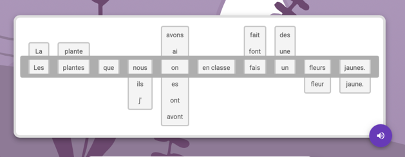}
\caption{Different positions of sliders creating a non-grammatical sentence (vector [2,2,1,3,3,3,1,1], see 4.3)}
\Description{Screenshot showing a sentence with incorrectly aligned sliders, creating a grammatically incorrect sentence in French.}
\label{fig:fig1}
\end{minipage}
\hfill
\begin{minipage}{0.48\textwidth}
\centering
\includegraphics[width=\textwidth]{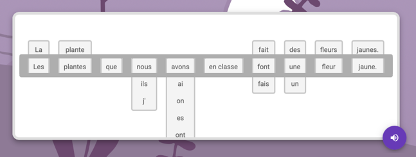}
\caption{Different positions of sliders creating a grammatical sentence (vector [2,2,1,1,2,2,2,2], see 4.3)}
\Description{Screenshot showing a sentence with correctly aligned sliders, creating a grammatically correct sentence in French.}
\label{fig:fig2}
\end{minipage}
\end{figure}

When the player considers a sentence may be correct, he or she can validate it and then gets feedback of correctness (green for correct, red for incorrect).

\begin{figure}[h]
\centering
\begin{minipage}{0.48\textwidth}
\centering
\includegraphics[width=\textwidth]{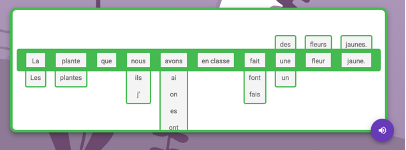}
\caption{What the player sees with positive (green) feedback}
\Description{Screenshot showing the game interface with green feedback indicating a grammatically correct sentence has been constructed.}
\label{fig:fig3}
\end{minipage}
\hfill
\begin{minipage}{0.48\textwidth}
\centering
\includegraphics[width=\textwidth]{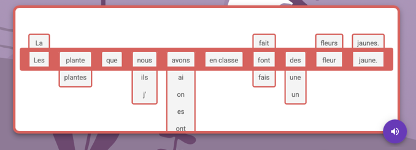}
\caption{What the player sees with negative (red) feedback}
\Description{Screenshot showing the game interface with red feedback indicating a grammatically incorrect sentence has been constructed.}
\label{fig:fig4}
\end{minipage}
\end{figure}

An important feature of the game is that it gives negative feedback without indicating the number of sliders that are incorrect, or which ones. This is a deliberate decision to force reflection, and we have always observed a progressive deeper attention of players with this gameplay.

Depending on the exercise and the syntactic structure, the number of sliders may be different. The actual distribution of exercises is shown in Table 1. Depending on the exercise and the words on each slider, the number of solutions may be different. In the present data set, the distribution is shown in Table 2.

\begin{table}[h]
\caption{Distribution of exercises per number of sliders}
\label{tab:sliders}
\centering
\begin{tabular}{cc}
\toprule
Number of sliders & Number of exercises \\
\midrule
2 & 3 \\
3 & 7 \\
4 & 2 \\
5 & 1 \\
6 & 2 \\
7 & 4 \\
8 & 9 \\
9 & 6 \\
10 & 4 \\
11 & 3 \\
12 & 1 \\
13 & 1 \\
\bottomrule
\end{tabular}
\end{table}

\begin{table}[h]
\caption{Distribution of exercises per number of solutions}
\label{tab:solutions}
\centering
\begin{tabular}{cc}
\toprule
Number of solutions & Number of exercises \\
\midrule
1 & 1 \\
2 & 6 \\
3 & 4 \\
4 & 7 \\
5 & 2 \\
6 & 8 \\
8 & 6 \\
9 & 1 \\
10 & 1 \\
12 & 1 \\
16 & 2 \\
18 & 3 \\
20 & 1 \\
\bottomrule
\end{tabular}
\end{table}

\subsection{Data}
In this paper, we concentrate the analyses on the period of April 2025 when the experiment ran smoothly in the different classrooms, without any access problem to the platform. During this period:
\begin{itemize}
\item 100 different students played at least once the game Tirettes (out of 185 students),
\item 25 different level games have been played (out of 43 Tirettes available),
\item 597 different sessions have been played.
\end{itemize}

The Adaptive Learning system was activated\footnote{The Adaptive Learning system uses the COMPER tools that allow a skills-based approach to support personalized learning through a competency framework (knowledge-based AI), the development of a skills profile for each learner, and a recommendation engine which organizes the next high-priority level games. https://comper.fr}, and the teachers could adapt settings according to their objectives and pedagogic intention (e.g. revision, consolidation, prerequisites, etc.). Moreover, to give learners more choice, they were always given three recommendations for the next exercise from the AI engine, with at least two different games. This meant that students could avoid the \textit{Tirettes} proposition if they wanted to. This explains why a large part of the students didn't play the \textit{Tirettes} and why only 25 levels were used.

Nevertheless, these 100 students played 597 sessions for which we have data including ID of the student, class, date, hour of beginning and ending (then duration), score, sequence of actions: movement of a slider, validation attempt, and the sentence validated (leading to a correct or incorrect sentence).

In these sessions, pupils attempted to validate 2,800 sentences. Since 143 of these were re-validations, we will analyze a total of 2,657 validations. Of these, 1,186 were grammatical sentences as expected (44.6\%), and 1,471 were non-grammatical (55.4\%). To achieve these 2,657 validations, 7,126 movements of sliders have been recorded, meaning a total of 9,783 actions over the 597 sessions.

\section{Results}
In this preliminary study, we concentrate on an overview of available data and preliminary analysis of tendencies, introducing the observable of convergence.

\subsection{General Results}
\begin{figure}[h]
\centering
\begin{minipage}{0.48\textwidth}
\centering
\includegraphics[width=\textwidth]{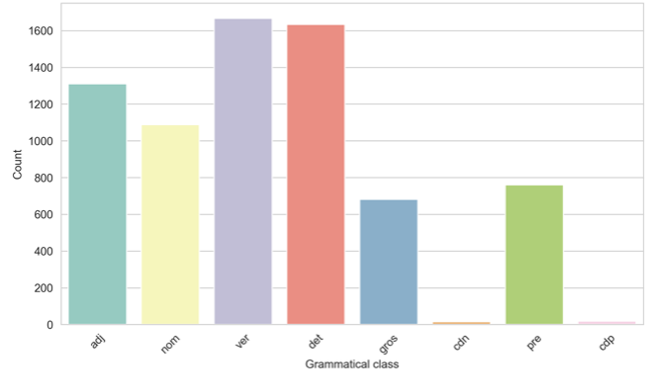}
\caption{General overview of results showing errors by grammatical category}
\Description{Bar chart showing the distribution of errors across different grammatical categories, with verbs and determiners having the highest error counts.}
\label{fig:fig5}
\end{minipage}
\hfill
\begin{minipage}{0.48\textwidth}
\centering
\includegraphics[width=\textwidth]{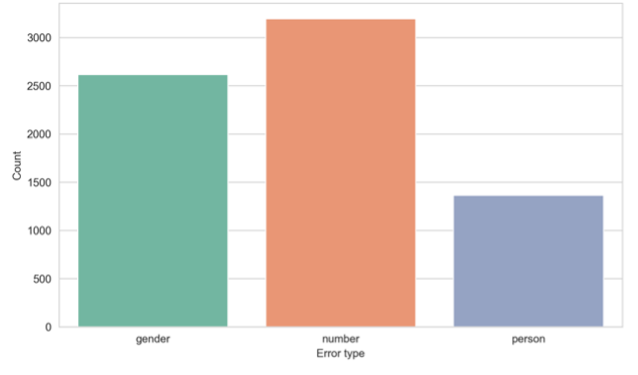}
\caption{Distribution of errors depending on grammatical characteristics (gender, number and person)}
\Description{Bar chart showing the distribution of errors across grammatical characteristics, with number errors being most prevalent, followed by gender and person errors.}
\label{fig:fig6}
\end{minipage}
\end{figure}

The verb is the grammatical category with the highest number of errors (1,665 errors). This is not surprising given the usual difficulties. However, errors in the use of determiners are almost at the same level (1,634 errors), indicating high levels of difficulty in choosing the correct determiner flexion. The fact that the game induces a building of the sentence that may not be linear (from left to right like writing) could explain high errors on a grammatical category that is usually less concerned. Next are the adjectives (1,310 errors) and nouns (1,088 errors). This distribution seems predictable, still according to the usual difficulties that can be observed in classical studies.

A majority of the errors breaking up the agreement chain are number errors (about 3,195), as would be expected, as this type of error can affect all types of chains (concerning noun, determiner, adjective, pronoun or verb). Then, we observe approximately 2,616 gender errors, which is significant as it primarily affects the agreement with the noun. Person errors are more limited (approximately 1,364), but they reflect the usual difficult agreement between the subject and the verb.

\begin{figure}[h]
\centering
\includegraphics[width=0.5\textwidth]{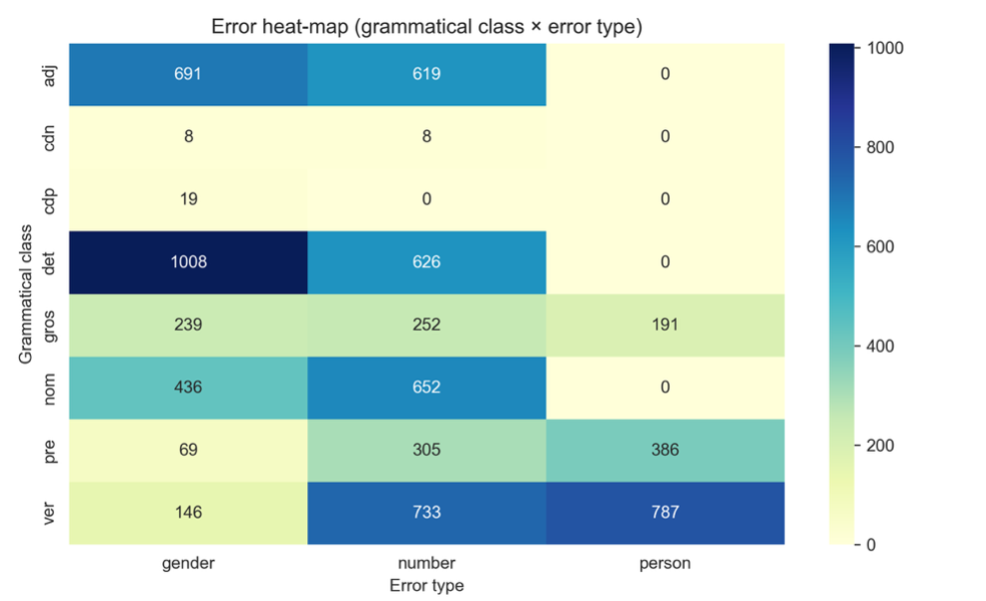}
\caption{Error heat-map (grammatical class × error type)}
\Description{Heat map displaying the distribution of errors across grammatical classes and error types, showing patterns of difficulty in different agreement contexts.}
\label{fig:fig7}
\end{figure}

Mixing these data, the error heat-map (Figure \ref{fig:fig7}) shows the distribution of errors according to grammatical category and grammatical features. When errors on adjective are almost equally linked to gender or number, this is very different with determiners where gender errors account for 61.7\% (1,008 errors of 1634). Considering the verb, number and person errors are the majority, probably due to the usual error consisting in choosing a verbal flexion ending with the morphogram -s instead of -ent for a plural, 3rd person. In French nouns and adjectives, -s is the usual plural morphogram but it indicates the 2nd person of singular in conjugation, leading to a double error according to verbal agreement (person and number); for example, a player could choose \textit{les enfants manges} instead of \textit{les enfants mangent} (the children eat, plural, 3rd person), with a double error as \textit{manges} is the verbal form for \textit{tu manges} (you eat, singular, 2nd person).

To go beyond this static overview, we need to consider actions and what happens before a student validates a built sentence.

\subsection{Tendencies: Overview}
A global account of 7,126 real sliders' moves has been recorded along the 597 sessions, meaning a mean of 11.94 sliders' moves per session.

Next two tables give the distribution of those moves per grammatical category (Table \ref{tab:moves_cat}) or syntactic function (Table \ref{tab:moves_func}), depending on the exercise. We keep 6,526 moves on sliders that may affect the structure and correctness of the sentence: determiner (\textit{det}), noun (\textit{nom}), adjective (\textit{adj}) and verb (\textit{ver}) for exercises with sliders indexing grammatical categories; and groups with subject function (GS) and group with predicate function (Pred) for exercises with sliders indexing syntactic functions.

\begin{table}[h]
\caption{Moves per grammatical category}
\label{tab:moves_cat}
\centering
\begin{tabular}{cc}
\toprule
Grammatical category & Total moves \\
\midrule
det & 1457 \\
nom & 1112 \\
adj & 916 \\
ver & 769 \\
total & 4254 \\
\bottomrule
\end{tabular}
\end{table}

\begin{table}[h]
\caption{Moves per syntactic grammatical functions}
\label{tab:moves_func}
\centering
\begin{tabular}{cc}
\toprule
Group function & Total moves \\
\midrule
GS & 1285 \\
Pred & 987 \\
total & 2272 \\
\bottomrule
\end{tabular}
\end{table}

As can be seen in Table \ref{tab:moves_cat}, it appears that the determiner's slider is the one that moves the most (1,457). As we saw previously (Figure \ref{fig:fig7}), the determiner is also one of the two grammatical categories with the most errors. This word, usually written in first in a chain agreement, seems difficult to agree in the case of the game.

The verb's slider moves the least (769), which may be unexpected. It is also the category with the most errors (Figure \ref{fig:fig5}). This could suggest that the verb is often primarily fixed before agreeing words forming the subject group. This strategy is the opposite of how we usually write and process words from left to right, so it may be difficult for students as it is unusual. This hypothesis also goes along with the difficulties observed on the determiner.

A similar result can be observed in exercises where the subject group and the predicate must be agreed (Table \ref{tab:moves_func}): the subject is more moved than the predicate (1285 moves vs 987). A result that may indicate again that the predicate is more often fixed first, and then there is an attempt to choose the subject form to fulfil the chain agreement.

These results suggest that we should explore the consequences of a slider's movement on the overall validity (or grammaticality) of a sentence. It can be done because each possible sentence is considered as a string of sliders' positions.

\subsection{Tendencies: The Hamming Distance}
Each exercise has several solutions (e.g. grammatically correct sentences), and the number of solutions can be very different (from 2 to 20, Table \ref{tab:solutions}). Each solution is equivalent to a specific vector (string of sliders' positions, see Figures 1 and 2). Considering a certain sliders' combination, it is then possible to calculate the difference between this vector and the closest vector that is a true solution, in the manner of a Hamming distance \cite{hamming1950}. We then call 'Hamming distance' the number of sliders' moves between a proposed sentence (seen as a string of sliders' positions) and the closest solution among the different solutions. A Hamming distance of 1 will then indicate that the player is one move from the solution (for example from vector [1,1,2] to vector [1,3,2]).

This allows now to focus on what happens after a slider's move. We have the comparison for 6124 moves (as the initial move m0 cannot be compared to any previous move m1). Table \ref{tab:hamming_cat} gives the impact of those moves on Hamming distance per grammatical category; table \ref{tab:hamming_func} gives the impact per syntactic function.

\begin{table}[h]
\caption{Impact of moves on Hamming distance per grammatical category}
\label{tab:hamming_cat}
\centering
\small
\begin{tabular}{lccc}
\toprule
Grammatical category & Improved (\%) & Worsened (\%) & Unchanged (\%) \\
\midrule
det & 33.0 & 44.9 & 22.1 \\
nom & 51.2 & 15.3 & 33.5 \\
adj & 40.8 & 23.6 & 35.6 \\
ver & 56.6 & 14.3 & 29.1 \\
\bottomrule
\end{tabular}
\end{table}

It appears that moves on the determiner slider worsen the situation more often than moves on the other sliders (Hamming distance is increased) (Table \ref{tab:hamming_cat}). Conversely, moves on the verb and noun sliders have a more positive impact (more than 50\% in both cases) and a reduced negative impact (14.3\% and 15.3\% respectively). There are fewer actions on the adjective with more dispersed consequences.

\begin{table}[h]
\caption{Impact of moves on Hamming distance per grammatical syntactic functions}
\label{tab:hamming_func}
\centering
\small
\begin{tabular}{lccc}
\toprule
Grammatical function & Improved (\%) & Worsened (\%) & Unchanged (\%) \\
\midrule
GS & 27.9 & 27.2 & 44.9 \\
pred & 50.7 & 17.3 & 32.0 \\
\bottomrule
\end{tabular}
\end{table}

When considering exercises with groups, actions on the predicate slider are more efficient than those on the subject slider (Table \ref{tab:hamming_func}). Hamming distance is improved half of the time, worsened only 17.3\% of the time.

However, we must consider a new variable: moving a slider may get the new vector closer to another solution (or gold-vector). In other terms, in case of several solutions (and probably more often in case of exercises with a high number of solutions), it is not obvious to know if a move leads to an improvement of the Hamming distance to the previous gold-vector, or to a new gold-vector. Study of the possibility of a gold-vector change following a move on a slider is synthesised in table \ref{tab:gold_cat} (per grammatical categories) and table \ref{tab:gold_func} (per syntactic functions).

\begin{table}[h]
\caption{Distribution gold-vector change following a move per grammatical category}
\label{tab:gold_cat}
\centering
\begin{tabular}{lcc}
\toprule
Grammatical category & Gold changed & Total actions \\
\midrule
det & 493 & 1457 \\
nom & 448 & 1112 \\
adj & 233 & 916 \\
ver & 115 & 769 \\
\bottomrule
\end{tabular}
\end{table}

\begin{table}[h]
\caption{Distribution gold-vector change following a move per syntactic grammatical function}
\label{tab:gold_func}
\centering
\begin{tabular}{lcc}
\toprule
Grammatical function & Gold changed & Total actions \\
\midrule
GS & 693 & 1271 \\
pred & 429 & 987 \\
\bottomrule
\end{tabular}
\end{table}

The low impact of a verb's slider move (Table \ref{tab:gold_cat}) can be explained by the usual few correct flexions available on the slider, and the link to several other words in the chain of agreement. This is confirmed by the highest impact of a change on the noun that is usually called "agreement giver" in grammatical analysis.

The difference of impact is very low for exercises with functions (Table \ref{tab:gold_func}) because learners have only two sliders, leading to a comparable probability.

At this stage of the study, the evolution of Hamming distance along time is the next step, with the question of the convergence towards a solution. Indeed, the game mechanics induce a sequence of actions (moves on sliders) to find a solution, i.e. a grammatically correct sentence where all words are agreed in their respective chains of agreement.

\subsection{Tendencies: The Convergence}
The next step in using the Hamming distance is to study how a student's work may converge towards a solution following the series of action on the slides.

\subsubsection{Overall Convergence}
\begin{figure}[h]
\centering
\includegraphics[width=0.5\textwidth]{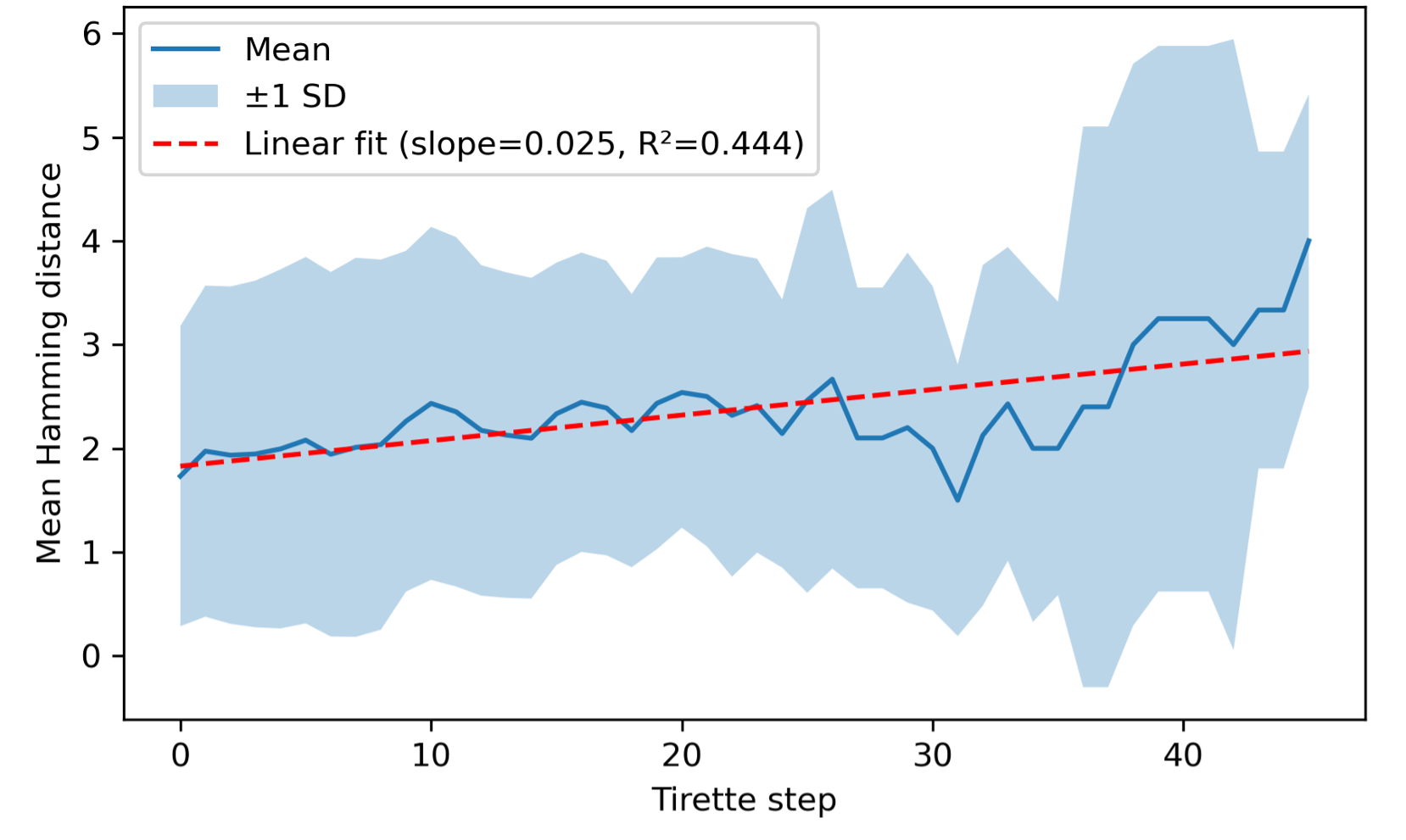}
\caption{Average convergence (all sessions) with linear trend}
\Description{Line graph showing the average Hamming distance over time across all sessions, with a positive linear trend indicating non-convergence on average.}
\label{fig:fig8}
\end{figure}

Figure \ref{fig:fig8} shows the average convergence for all sessions, showing a very irregular progression and a positive slope for linear trend indicating an absence of convergence. However, this first impression must be qualified: this figure is the mean tendency of 100 students playing several exercises of very different difficulties, then a general approach cannot be conclusive. A few general trends can be identified:
\begin{itemize}
\item The standard deviation shows a large discrepancy that reduces slightly around step 20, before increasing again due to the small number of students who continue to struggle.
\item The linear trend is undoubtedly disrupted by some students who take a very high number of actions to find a solution. Although there are few of them, they influence the right part of the distribution. A look at the data confirms this: few students with Hamming distance that can be very high (see next section).
\item There may be a conflict between trends for exercises using grammatical categories and grammatical functions, between trends for different chains of agreement.
\end{itemize}

In short, the presence of too many variables means that general average convergence is not an efficient analysis tool.

Then, a better approach could be to consider only chains including the verb, thus the principal chain of a sentence, excluding (temporarily) other chains and exercises with two sliders for two groups (subject and predicate).

\subsubsection{Convergence for Main Chain (Including Verb)}
\begin{figure}[h]
\centering
\includegraphics[width=0.5\textwidth]{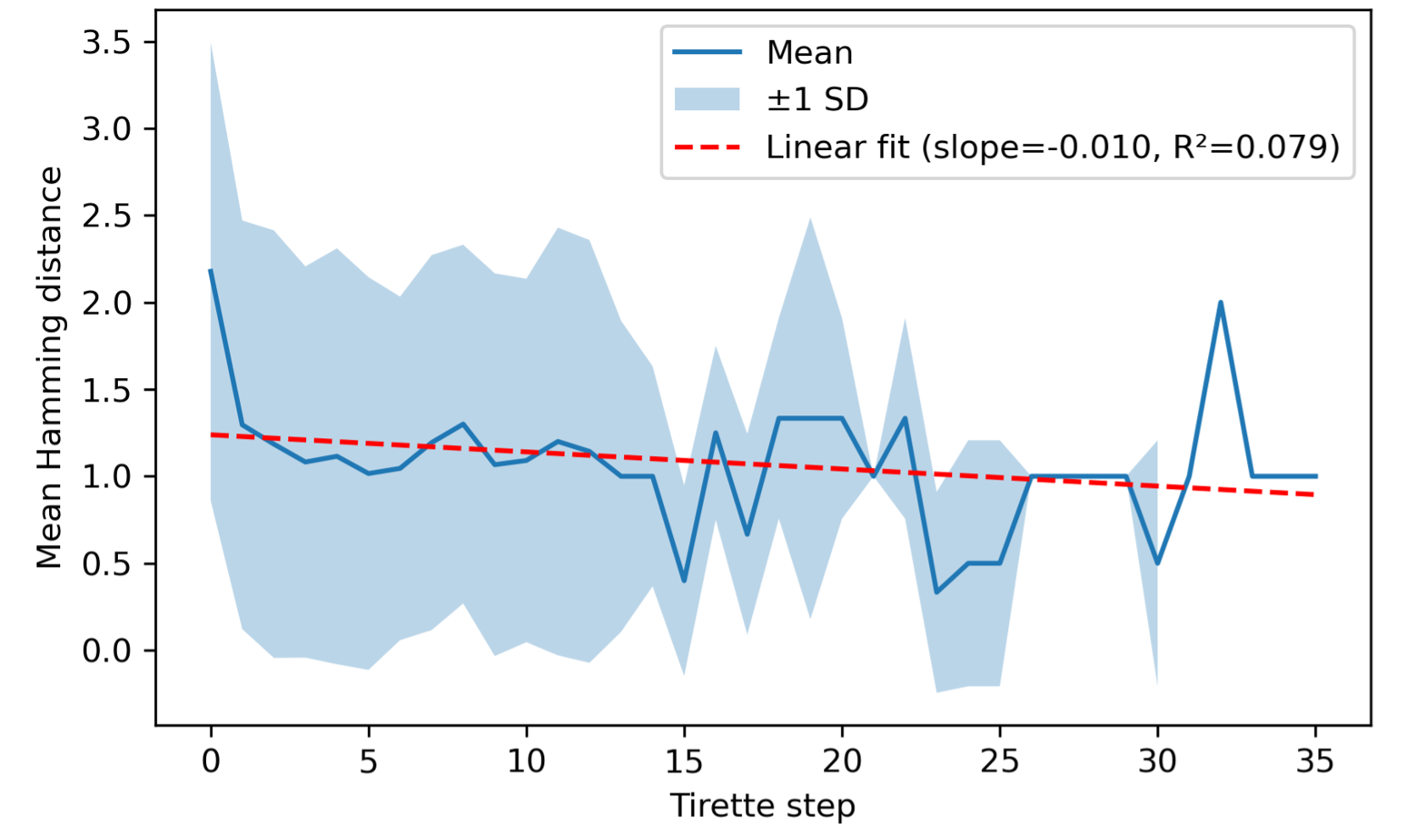}
\caption{Average convergence (all sessions) with linear trend for chain including verb}
\Description{Line graph showing the average Hamming distance over time across all sessions focusing on chains that include verbs, with a negative linear trend indicating convergence.}
\label{fig:fig9}
\end{figure}

The new average convergence for all sessions is showed in Figure \ref{fig:fig9}. This time the slope is negative, the trend is towards convergence. Several very negative spikes can be observed (at the beginning, first 2 moves; then at 15, 23 and 30 steps) indicating several convergences, probably corresponding to different students' activity and different exercises of various difficulty. Once again, the end of the curve is due to very few students that can't achieve the game.

Once more, it is possible to take it a step further and analyse the convergence for some very representative exercises.

\subsubsection{Convergence for Specific Exercises}
We have decided to study four exercises for which the general convergence trend seemed interesting.

\begin{figure}[h]
\centering
\begin{minipage}{0.48\textwidth}
\centering
\includegraphics[width=\textwidth]{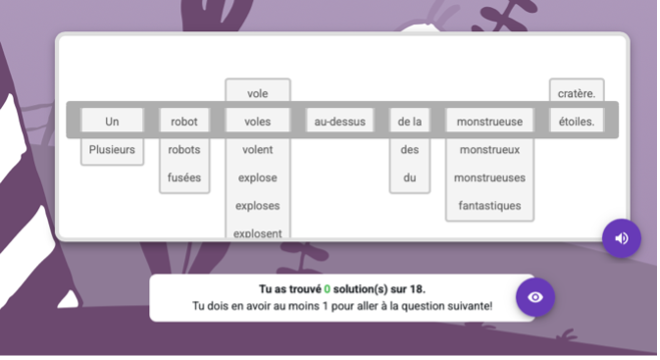}
\caption{Screenshot of exercise XP6H\_11}
\Description{Screenshot of a game exercise with 6 sliders where the player must construct a grammatically correct French sentence by aligning word forms.}
\label{fig:fig10}
\end{minipage}
\hfill
\begin{minipage}{0.48\textwidth}
\centering
\includegraphics[width=\textwidth]{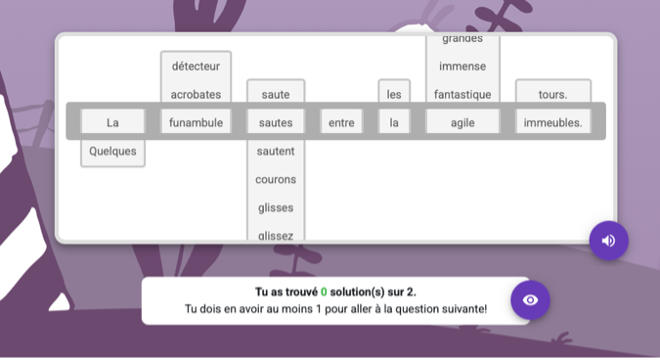}
\caption{Screenshot of exercise XP6H\_12}
\Description{Screenshot of a game exercise with 6 sliders where the player must construct a grammatically correct French sentence by aligning word forms.}
\label{fig:fig11}
\end{minipage}
\end{figure}

\begin{figure}[h]
\centering
\begin{minipage}{0.48\textwidth}
\centering
\includegraphics[width=\textwidth]{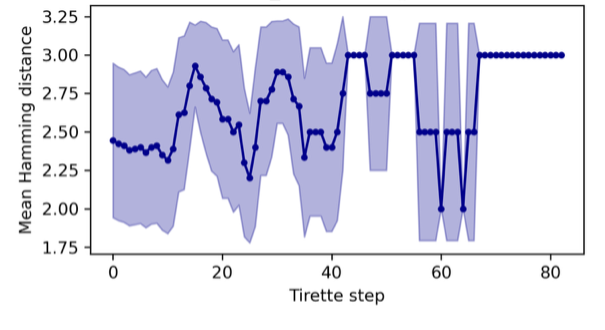}
\caption{Average convergence exercise XP6H\_11}
\Description{Line graph showing the average Hamming distance over time for exercise XP6H\_11, showing relatively stable convergence patterns due to multiple possible solutions.}
\label{fig:fig12}
\end{minipage}
\hfill
\begin{minipage}{0.48\textwidth}
\centering
\includegraphics[width=\textwidth]{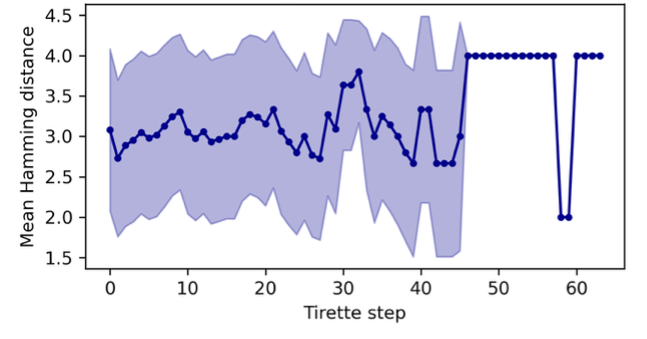}
\caption{Average convergence exercise XP6H\_12}
\Description{Line graph showing the average Hamming distance over time for exercise XP6H\_12, showing more erratic convergence patterns due to fewer possible solutions.}
\label{fig:fig13}
\end{minipage}
\end{figure}

Exercises XP6H\_11 and XP6H\_12 are the same in their structure (6 sliders, Figures \ref{fig:fig10} and \ref{fig:fig11}) but one has 18 solutions (XP6H\_11) when the other has only 2 solutions (Table \ref{tab:exercises}). This leads to quite different average convergence profiles (Figures \ref{fig:fig12} and \ref{fig:fig13}). They have respectively been played by 40 and 54 students for 295 and 413 actions (Table \ref{tab:exercises_actions}).

Exercise 7H\_6 has also 6 sliders, but with 6 solutions (Table \ref{tab:exercises}). It has been played by 78 students for 408 actions.

\begin{figure}[h]
\centering
\begin{minipage}{0.48\textwidth}
\centering
\includegraphics[width=\textwidth]{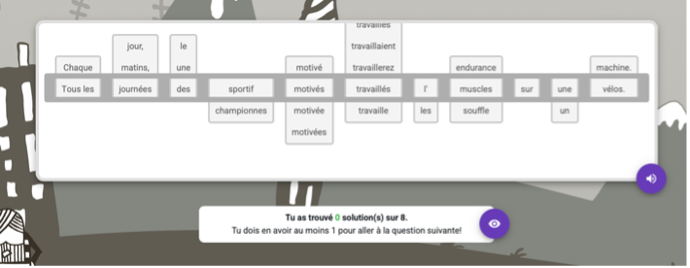}
\caption{Screenshot of exercise XP6H\_22}
\Description{Screenshot of a game exercise with 10 sliders where the player must construct a grammatically correct French sentence by aligning word forms.}
\label{fig:fig14}
\end{minipage}
\hfill
\begin{minipage}{0.48\textwidth}
\centering
\includegraphics[width=\textwidth]{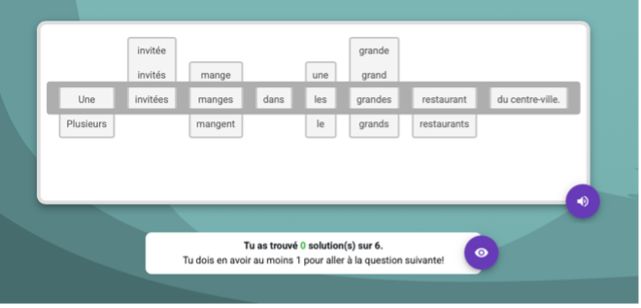}
\caption{Screenshot of exercise XP7H\_6}
\Description{Screenshot of a game exercise with 6 sliders where the player must construct a grammatically correct French sentence by aligning word forms.}
\label{fig:fig15}
\end{minipage}
\end{figure}

Exercise 6H\_22 is very different: 10 sliders with 8 solutions. It has been played only by 2 students for 45 actions.

\begin{table}[h]
\caption{Number of sliders and solutions per exercise}
\label{tab:exercises}
\centering
\begin{tabular}{lcc}
\toprule
ID Exercise & Number of sliders & Number of solutions \\
\midrule
XP6H\_11 & 6 & 18 \\
XP6H\_12 & 6 & 2 \\
XP6H\_22 & 10 & 8 \\
XP7H\_6 & 6 & 6 \\
\bottomrule
\end{tabular}
\end{table}

\begin{table}[h]
\caption{Number of players and actions per exercise}
\label{tab:exercises_actions}
\small
\centering
\begin{tabular*}{\columnwidth}{@{\extracolsep{\fill}}lrrr@{}}
\toprule
ID Exercise & \multicolumn{1}{c}{Players} & \multicolumn{1}{c}{Slider moves} & \multicolumn{1}{c}{Validations} \\
& \multicolumn{1}{c}{(count)} & \multicolumn{1}{c}{(count)} & \multicolumn{1}{c}{(count)} \\
\midrule
XP6H\_11 & 40 & 782 & 295 \\
XP6H\_12 & 54 & 1103 & 413 \\
XP6H\_22 & 2 & 107 & 45 \\
XP7H\_6 & 78 & 1226 & 408 \\
\bottomrule
\end{tabular*}
\end{table}

\begin{figure}[h]
\centering
\begin{minipage}{0.48\textwidth}
\centering
\includegraphics[width=\textwidth]{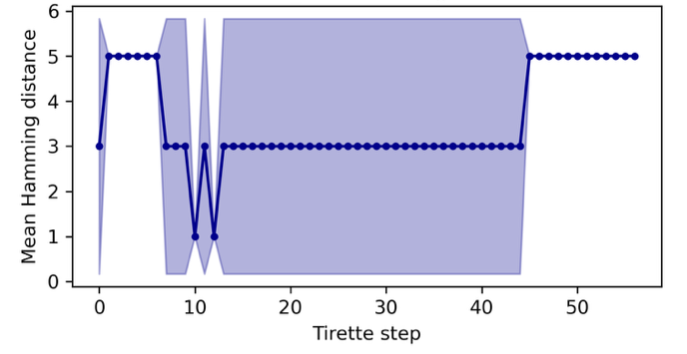}
\caption{Average convergence exercise XP6H\_22}
\Description{Line graph showing the average Hamming distance over time for exercise XP6H\_22, with significant fluctuations due to the complexity of the exercise.}
\label{fig:fig16}
\end{minipage}
\hfill
\begin{minipage}{0.48\textwidth}
\centering
\includegraphics[width=\textwidth]{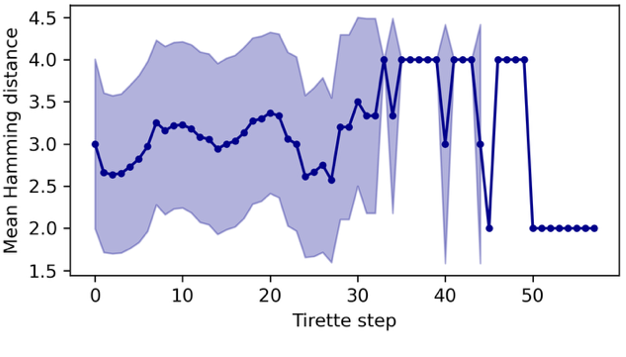}
\caption{Average convergence exercise XP7H\_6}
\Description{Line graph showing the average Hamming distance over time for exercise XP7H\_6, showing a gradual convergence pattern with moderate fluctuations.}
\label{fig:fig17}
\end{minipage}
\end{figure}

These 4 exercises illustrate the large variety among exercises, with a lot of variables, and the adaptive learning recommendation system (as well as objectives defined in classrooms) induce very different use of them. All these variables make it difficult to give general trends. Nevertheless, in this case, average convergences appear different. The profile of XP6H\_11 is quite flat, certainly indicating that students progressively succeed without getting far from solutions because there are many solutions. XP6H\_12, although with the same syntactic structure, shows a more disrupted average convergence trend, certainly indicating that more students tend to be further from a solution. Standard deviation is clearly higher here.

XP7H\_6, with a similar structure than the two previous games, and with 6 solutions, has a profile that could be linked to XP6H\_12, with higher variations, higher standard deviation. The lack of solutions is a heavy factor of difficulty.

\begin{figure}[h]
\centering
\begin{minipage}{0.48\textwidth}
\centering
\includegraphics[width=\textwidth]{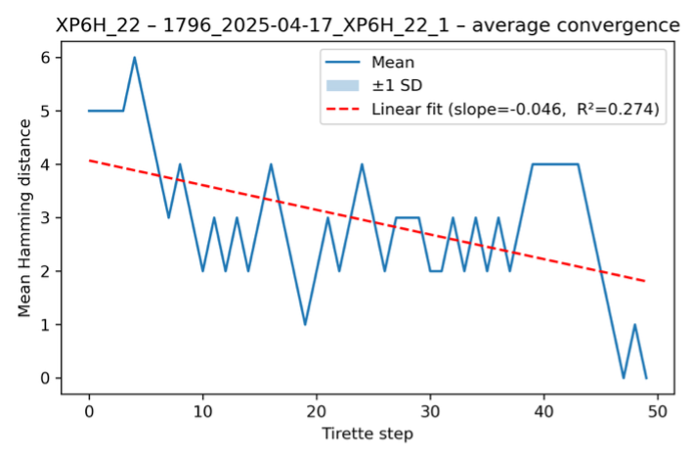}
\caption{Average convergence, XP6H\_22, student 1796}
\Description{Line graph showing the Hamming distance trajectory for student 1796 on exercise XP6H\_22, showing a progressive convergence with fluctuations.}
\label{fig:fig18}
\end{minipage}
\hfill
\begin{minipage}{0.48\textwidth}
\centering
\includegraphics[width=\textwidth]{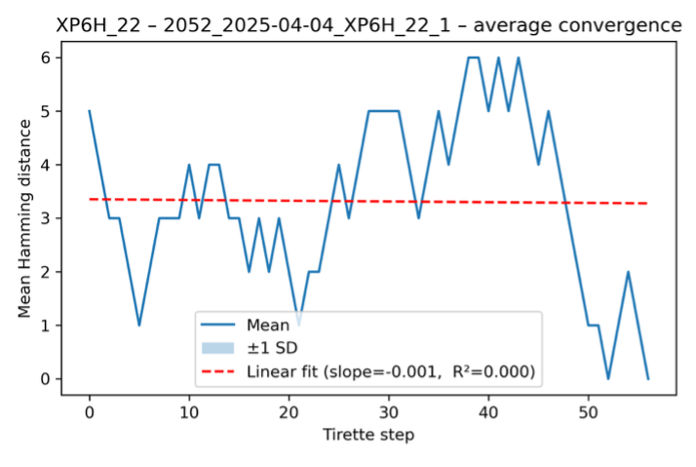}
\caption{Average convergence, XP6H\_22, student 2052}
\Description{Line graph showing the Hamming distance trajectory for student 2052 on exercise XP6H\_22, showing an erratic pattern with periods of being far from solutions.}
\label{fig:fig19}
\end{minipage}
\end{figure}

Finally, the game XP6H\_22 is very different and has been played only twice, leading to an average convergence that is difficult to read. We use this exercise to go deeper in data and compare the convergence for the two students (Figures \ref{fig:fig18} and \ref{fig:fig19}).

Student 1796 has a progressive convergence towards a solution (slope is negative) with lots of variations of +/-1 in Hamming distance during a series of actions. It is possible to see that he or she comes closer to a solution a first time, then moving further to finally find a solution. Student 2052's average convergence is close to zero, but the progression is slightly different with two actions reducing Hamming distance to 1 at the beginning, then a long series of moves far from any solutions before moving quickly to a solution.

A closer look at the data depending on grammatical categories enlighten these differences: student 1796 never moves the verb slider, and very rarely the adjective slider. Everything is like struggling with determiner and noun before finding a solution of agreement. In the meantime, student 2052 acts on each slider, rarely on adjective, mostly on determiner and noun although they wear a few items only.

What seems to appear here is the high difficulty, for these two students, to deal with the determiner-noun agreement, a quite surprising observation. The other main point is the interest of the Hamming distance as a direct indicator of the convergence (or not). The contribution is rich as it allows a real-time monitoring of the work of each student.

\section{Discussion and Perspectives}
We conducted the analyses of the data coming from the use in 11 primary school classrooms of a game-based learning platform with adaptive learning, focusing on a specific game that requires to build a grammatically correct sentence in French acting on sliders, each slider wearing several flexions of a word and/or several words from the same grammatical category. The specific mechanics of this game reveal the process of grammatically matching the words in a sentence, resulting in two main actions: moving the sliders and validating attempts. All these actions are collected to form the available database.

A first output from this database has been to give teachers access to a "replayer" that gives the possibility to look at the series of actions of a specific player on a specific game, during a specific session\footnote{\url{https://files.atypon.com/acm/d311b7857b8e37a1affeb136fadff1ba}}. This tool is a real innovation brought to primary school teachers. Actual actions of students are usually non accessible to the teacher. Using the data collected makes it possible and replayers are currently available for most games on the platform. However, this paper also delivers an in-depth analysis of the data, using learning analytics to decipher children grammatical reasoning when playing the interactive grammatical game.

We find that the high number of variables (huge diversity of exercises inside the game; diversity of students and diversity of working conditions as learning pathways were personalized by the adaptive learning system) make the overall views impossible to exploit and analyse directly. Descriptive statistics are not helpful here. However, considering the series of actions and trying to give a modelisation brings new horizons. The Hamming distance which measures the gap between a combination of sliders (i.e. a vector) that is proposed by a student at a given moment, and a solution (i.e. a correct combination or 'gold vector') is important for understanding what happens when a slider is moved and for analysing the evolution over time and possible convergence towards a solution. Then we can search patterns that would enlighten strategies.

Once again, the results are far more relevant when fine-grained data is used. We can evaluate the consequence of a slider's move in terms of distance to the closest solution. And we can detect if this consequence is different depending on the index of the slider (grammatical category or function, and which one?). With this scope, we can highlight a few points:
\begin{itemize}
\item The verb seems to be chosen before working on other sliders. The sequence of actions tends towards attempts of agreement of the determiner, noun and adjective after the verb. This is aligned with what can be taught in French speaking classrooms where the initial identification of the verb is recommended. Then, the students seem to struggle with the agreement of the following words. This requires them to go to the left of the sentence, rather than working from left to right as they would when writing. The unusually high level of actions and errors on the determiner seems to illustrate this.
\item Average convergence graphs become unreadable over time because there are a few students left, apparently moving sliders without any agreement strategy: rather trial and error, or random attempts.
\item Number of sliders appears to be an obvious criterion of difficulty of an exercise compared to another.
\item Above all, the number of solutions is very impactful (see end of section 4.4) and must be highly considered when creating a new exercise. Indeed, adding or removing one or two words on a slider may change the total number of solutions and then the way a student will be able to converge towards at least one solution.
\item The individual convergence trends are very useful as they enlighten sequences that bring the student closer to or further away from a solution, including considering the a possible change in the closest gold-vector.
\end{itemize}

This last point is essential to us; it is the gateway to real-time scaffolding and stimulus. Indeed, we could imagine several scenarios exploiting an interactive and adaptive interface using learning analytics. The following scenarios are hypothetic interactions based on what a teacher may decide in a differentiated teaching process in the classroom.
\begin{enumerate}
\item The student is converging towards a solution then moves away:
\begin{itemize}
\item A helper is sent (via an avatar for example) to warn him or her, without any further feedback.
\item Additional feedback could be the proposal to come back to the closest previous combination.
\item Another feedback could be a reminder of a lesson (for example about the determiner and how it matches with a noun according to gender and number).
\end{itemize}
\item The student is far from a solution:
\begin{itemize}
\item A helper is sent to warn him or her.
\item Additional feedback could be the proposal to restart the exercise.
\item Another feedback could be the proposal to provide a clue (such as fixing one slider correctly).
\end{itemize}
\item The student doesn't use a strategy that has been found efficient through patterns' analysis:
\begin{itemize}
\item A helper is sent to warn him or her.
\item Additional feedback could be an explanation or an advice depending on the gold-strategy (fixing verb first, or noun, etc.)
\end{itemize}
\item The student doesn't use additional features that usually help him or her (text-to-speech or opendyslexic font, for example) \cite{dherbey2021}:
\begin{itemize}
\item A helper is sent to warn him or her.
\item Additional feedback could be an animation to remind how to activate it.
\end{itemize}
\item The student seems to be unexpectedly inactive or, on the contrary, validating randomly, too fast, etc.:
\begin{itemize}
\item A helper is sent to warn him or her.
\item Additional feedback could give advice on how to better achieve an exercise.
\end{itemize}
\end{enumerate}

In this set of scenarios, the first three are linked to real-time activity and strategy, and are then linked to learning or performing. They are similar to personalised scaffolding \cite{bruner1983} but based on learning analytics. Scenarios 4 and 5 differ in that they focus more on maintaining student engagement, like the teacher's usual interactive classroom regulation \cite{roogiers2003}. The next step is to develop the functionalities based on fine-grained learning analytics, implement them, and document their relevance and efficiency in supporting inclusive teaching and learning in the classroom.

This paper presents an initial analysis of a unique dataset collected in French speaking primary classrooms via an AI-based learning platform. Our analysis focused on learning analytics to understand students' grammatical reasoning strategies and to outline future developments in real-time scaffolding and interactive regulation. A next paper will deepen sequence mining and will test mathematical models to validate inferred strategies.

In addition to this analysis, we would like to highlight the potential of new tools for teachers. The replayer is a real innovation that provides access to students' actual actions. This is a time delayed use of the data, supporting the retroactive management of students and certainly helping differentiation. Real-time activity and strategy analysis will provide teachers with additional information, offering real-time feedback on classroom activities and summarising what occurred during a session. This feedback should help teachers to use the replayer to provide better individual monitoring and support for their students.

\begin{acks}
This research was supported by the Swiss National Science Foundation (SNSF) under grant n°100019\_215373 in the framework of the DOCTA\textsuperscript{2}LE-FR project. We would like to thank the teachers and students who participated in this study, as well as the technical and research team who developed the GamesHub platform.
\end{acks}

\bibliographystyle{ACM-Reference-Format}
\bibliography{references}

\end{document}